\documentclass{article}

\usepackage{arxiv}

\usepackage[utf8]{inputenc} % allow utf-8 input
\usepackage[T1]{fontenc}    % use 8-bit T1 fonts
\usepackage{hyperref}       % hyperlinks
\usepackage{url}            % simple URL typesetting
\usepackage{booktabs}       % professional-quality tables
\usepackage{amsfonts}       % blackboard math symbols
\usepackage{nicefrac}       % compact symbols for 1/2, etc.
\usepackage{microtype}      % microtypography
\usepackage{lipsum}         % Can be removed after putting your text content
\usepackage{graphicx}
\usepackage{doi}
\usepackage{amssymb}

\usepackage{algorithm}
\usepackage{xcolor}
\usepackage{amsmath,amsthm,amssymb,amsfonts,bm}
\usepackage[]{algpseudocode}
\usepackage{makecell}
\usepackage{bbm}

\makeatletter
\newcommand*{\rom}[1]{\expandafter\@slowromancap\romannumeral #1@}
\DeclareMathOperator*{\tsum}{\textstyle\sum}

\newtheorem{theorem}{Theorem}

\usepackage{cleveref}

\renewcommand{\v}[1]{\ensuremath{\boldsymbol{#1}}}

\title{From Centralized to Decentralized Federated Learning: Theoretical Insights, Privacy Preservation, and Robustness Challenges}

\date{}
\author{ Qiongxiu Li\\
	Aalborg University\\
    Denmark \\
	\texttt{qili@es.aau.dk} \\
	\And
	Wenrui Yu\\
	Aalborg University\\
    Denmark \\
	\texttt{wenyu@es.aau.dk} \\
    \And
    Yufei Xia\\
	IP Paris \\
    France \\
	\texttt{yufei.xia@ip-paris.fr} \\
	\And
	Jun Pang\\
	University of Luxembourg\\  Luxembourg \\
	\texttt{jun.pang@uni.lu} 
}
\renewcommand{\headeright}{}
\renewcommand{\undertitle}{}
\hypersetup{
pdftitle={From Centralized to Decentralized Federated Learning: Theory, Methods and Challenges},
pdfsubject={},
pdfauthor={Qiongxiu Li, Wenrui Yu, Yufei Xia, Jun Pang},
pdfkeywords={Federated Learning, decentralized learning, privacy and security, robustness},
}

\begin{document}
\maketitle

\begin{abstract}
Federated Learning (FL) enables collaborative learning without directly sharing individual's raw data. FL can be implemented in either a centralized (server-based) or decentralized (peer-to-peer) manner. In this survey, we present a novel perspective: the fundamental difference between \emph{centralized} FL (CFL) and \emph{decentralized} FL (DFL) is not merely the network topology, but the underlying training protocol: \textit{separate aggregation vs. joint optimization}. We argue that this distinction in protocol leads to significant differences in model utility, privacy preservation, and robustness to attacks. We systematically review and categorize existing works in both CFL and DFL according to the type of protocol they employ. This taxonomy provides deeper insights into prior research and clarifies how various approaches relate or differ. Through our analysis, we identify key gaps in the literature. In particular, we observe a surprising lack of exploration of DFL approaches based on distributed optimization methods, despite their potential advantages. We highlight this under-explored direction and call for more research on leveraging distributed optimization for federated learning. Overall, this work offers a comprehensive overview from centralized to decentralized FL, sheds new light on the core distinctions between approaches, and outlines open challenges and future directions for the field.
\end{abstract}

% keywords can be removed

\section{Introduction}
Federated Learning (FL) is a distributed machine learning paradigm that enables multiple participants (nodes or clients) to collaboratively train a global model without directly sharing their raw data~\cite{mcmahan2017communication,li2020federated2}. In a typical FL setting, each client computes updates (e.g., gradients or model weights) from its local data and shares only those updates for aggregation. Two network topologies are commonly used for FL: the conventional centralized (or star) topology and the fully decentralized topology, as illustrated in Figure~\ref{fig:topo}.

In the \textit{centralized FL} (CFL) architecture, a central server coordinates the training process by interacting with all clients. The server collects model updates from clients, aggregates them into a global model, and then distributes the global model back to the clients. Clients periodically incorporate the global model into their local training and send updated models back to the server, repeating until convergence. This centralized protocol (often referred to as FedAvg \cite{mcmahan2017communication}) has been widely adopted in various applications \cite{ma2022assisted,cui2020personalized,jiang2020federated,yuan2022fedstn}. However, maintaining a central server in FL can be problematic. The server becomes a communication bottleneck and a single point of failure for the entire network \cite{zhang2020federated,hu2019decentralized}. All clients must trust the server to correctly aggregate updates and safeguard the integrity of the global model. If the server is compromised or goes offline, the learning process is disrupted. Moreover, a malicious server or a successful attack on the server could corrupt the global model or leak sensitive information \cite{security2023pasquini}. These vulnerabilities motivate the need for an alternative approach.

\textit{Decentralized FL} (DFL) eliminates the central server by enabling clients to cooperate in a peer-to-peer manner~\cite{hu2019decentralized,lian2017can,koloskova2020unified,sun2022decentralized,niwa2020edge,yu2024provable}. Instead of sending updates to a single server, each client exchanges model information with a subset of other clients (its neighbors) over a communication graph. Through iterative local communications, the network’s collective behavior approximates global aggregation dynamics, driving consensus toward a unified global model. Common aggregation approaches include gossiping SGD~\cite{jin2016scale}, D-PSGD~\cite{lian2017can}, and other variations ~\cite{tang2018d,hu2019decentralized}. By removing the central coordinator, DFL offers enhanced robustness (no single point of failure) and potentially better scalability in heterogeneous or dynamic network environments. It also mitigates the need to fully trust any single entity with the entire training process, since model aggregation is inherently distributed among the participants. On the other hand, decentralization introduces challenges such as slower convergence compared to centralized FL and increased complexity in managing communication overhead or inconsistent connectivity~ \cite{wang2024decentralized}.

Existing work often categorizes FL approaches by their centralization degree or graph topology. In this paper, we argue that the choice of communication or aggregation protocol is equally critical.   A typical FL protocol involves three sequential steps: 1) global model initialization, 2) local model updates, and 3) aggregation of updates to refine the global model. However, some frameworks unify steps 2 and 3 through distributed optimization algorithms that solve network-constrained problems directly, independent of topology. These include methods like the Alternating Direction Method of Multipliers (ADMM)  \cite{boyd2011distributed} and the Primal-Dual Method of Multipliers (PDMM)  \cite{zhang2018distributed,sherson2018derivation}.  To formalize this distinction, we classify the first paradigm as separated aggregation (sequential local training and aggregation) and the second as joint optimization (learning with embedded topology constraints).   This taxonomy highlights how protocol design, not just topology, shapes convergence, privacy, and communication efficiency in FL systems.

In this paper, we bridge CFL and DFL methods under a unified perspective. We systematically categorize prior work along two axes: (i) the presence of a central coordinator (CFL vs. DFL) and (ii) the use of seperate aggregation versus joint optimization algorithms. This dual framework positions existing algorithms in a clear taxonomy  (as summarized in Table~\ref{table:fl_form}), highlighting commonalities and differences that were previously obscured when considering topology alone. Through this analysis, several insights emerge. Notably, we find that most so-called “decentralized” FL methods follow the separate-aggregation paradigm (using, for example, peer-to-peer averaging protocols that mimic FedAvg), and the other joint optimization-based DFL strategies are relatively under-explored in practice. Our survey of the literature reveals a gap in applying advanced distributed optimization techniques to FL, despite their potential to improve efficiency, privacy, and robustness. 

The key contributions of this work are summarized as follows:
\begin{itemize}
    \item \textbf{New perspective on CFL vs. DFL:} In addition to the conventional view that the primary difference between centralized and DFL is the network topology. We argue that the \emph{underlying protocol} (separate aggregation vs. joint optimization) is the fundamental differentiator. We demonstrate how this perspective leads to a better understanding of performance trade-offs in utility (model accuracy and convergence), privacy, and robustness to various attacks.
    \item \textbf{Systematic literature review on FL's privacy and robustness:} We conduct a thorough review of existing FL research, spanning both CFL and DFL. We categorize these works based on several creatia including 1) centralized or decentralized architectures, 2) deployed training protocols 3) performances against passive and active adversaries. By organizing prior approaches into this taxonomy, we provide clarity on how various methods relate, and we illuminate the landscape of FL research in a way that aids practitioners and researchers in identifying appropriate techniques for their needs.
    \item \textbf{Identification of research gaps and future directions:} Based on our review, we pinpoint critical open challenges. In particular, we observe that \emph{DFL with fully distributed optimization algorithms} (as opposed to heuristic averaging) is an under-explored area with significant potential benefits. Few studies have leveraged advanced distributed optimization methods in FL, and we advocate for their broader adoption. We also highlight unresolved security and privacy issues, such as the need for provably robust protocols in both CFL and DFL settings.
\end{itemize}

% In this paper, we provide a comprehensive overview of the state-of-the-art in both centralized and decentralized federated learning, with a particular focus on the transition towards decentralization. We first review the fundamental concepts and notation used in FL (Section~\ref{}) and outline the standard centralized FL workflow along with its recent enhancements (Section~\ref{}). We then delve into DFL methods (Section~\ref{}), categorizing them into two broad classes: (i) \textit{average-consensus-based protocols}, which rely on iterative averaging (e.g., gossip algorithms [43]) to propagate model updates, and (ii) \textit{optimization-based protocols}, which formulate training as a constrained optimization problem solved by distributed algorithms like ADMM~\cite{mota2013d,li2019communication,chen2021coded} or PDMM~\cite{zhang2017distributed,sherson2018derivation,niwa2020edge}. Next, we examine the security and privacy aspects of FL. 

The remainder of this paper is structured as follows. In Section~\ref{sec.pre}, we provide necessary backgrounds such as notation, network setup, and threat models. Section~\ref{sec.cfl} reviews CFL methods and analyzes their aggregation protocols. Section~\ref{sec.dfl}  summarizes DFL approaches and relates them to CFL counterparts. In Section~\ref{sec.passive}, we compare CFL and DFL under passive adversary settings, evaluating their resilience to privacy attacks (e.g., membership inference, model inversion) and corresponding defense mechanisms. Sections~\ref{sec.active} and ~\ref{sec.robust} shift focus to active adversaries by first analyzing attack strategies (e.g., Byzantine attacks, data poisoning), and then reviewing defense mechanisms. Finally, Section~\ref{sec.conclusion} synthesizes key insights, identifies open challenges, and outlines future research directions for advancing FL efficiency, privacy,  and robustness.

% DFL protocols, also known as peer-to-peer learning protocols, fall into two main categories. The first involves average-consensus-based protocols. With these protocols, instead of sending model parameters to a central server, nodes collaborate together to perform model aggregation nodes in a distributed manner. The aggregation is typically done by partially averaging the local updates within a node's neighborhood.  Examples of these protocols are the empirical methods where the aggregation is done using average consensus techniques such as gossiping SGD~\cite{jin2016scale}, D-PSGD~\cite{lian2017can}, and variations thereof~\cite{tang2018d,hu2019decentralized}. 
%  The second category comprises protocols that are based on distributed optimization, referred to as optimization-based DFL. These (iterative) methods %do not separate the steps of local model training and model aggregation, but 
% directly formulate the underlying problem as a constrained optimization problem and employ distributed solvers like ADMM~\cite{mota2013d,li2019communication,chen2021coded} or PDMM~\cite{zhang2017distributed,sherson2018derivation,niwa2020edge} to solve them. The constraints are formulated in such a way that, upon convergence, the learned models at all nodes are identical. Hence, there is no explicit separation between updating local models and the update of the global model, i.e., the three steps in centralized FL mentioned before are executed simultaneously. 

\section{Preliminaries}\label{sec.pre}
In this section, we review key fundamentals that are necessary for understanding the rest of the paper.

\subsection{Notation}
Let $\mathbb{R}^d$ denote $d$-dimensional real vectors and $\mathbb{R}^{m \times n}$ the space of $m \times n$ real matrices. The symbols $\succ, \succeq, \prec$ and $\preceq$ denote generalised inequalities; between vectors it represents component-wise inequality.  Calligraphic letters denote sets and the Euclidean norm $\|\mathbf{x}\|$ for $\mathbf{x} \in \mathbb{R}^d$ derives from the inner product $\mathbf{x}^\intercal\mathbf{x}$.  Iterative processes are tracked via superscripts (e.g., $\mathbf{w}^{(t)}$ for model parameters $\mathbf{w}$ at iteration $t$). Consider $x$ as the realization of a random variable, the corresponding random variable will be denoted by the corresponding capital, i.e., $X$. 

\subsection{Network setup}
\begin{figure}[ht]
    \centering
\includegraphics[width=0.7\textwidth]{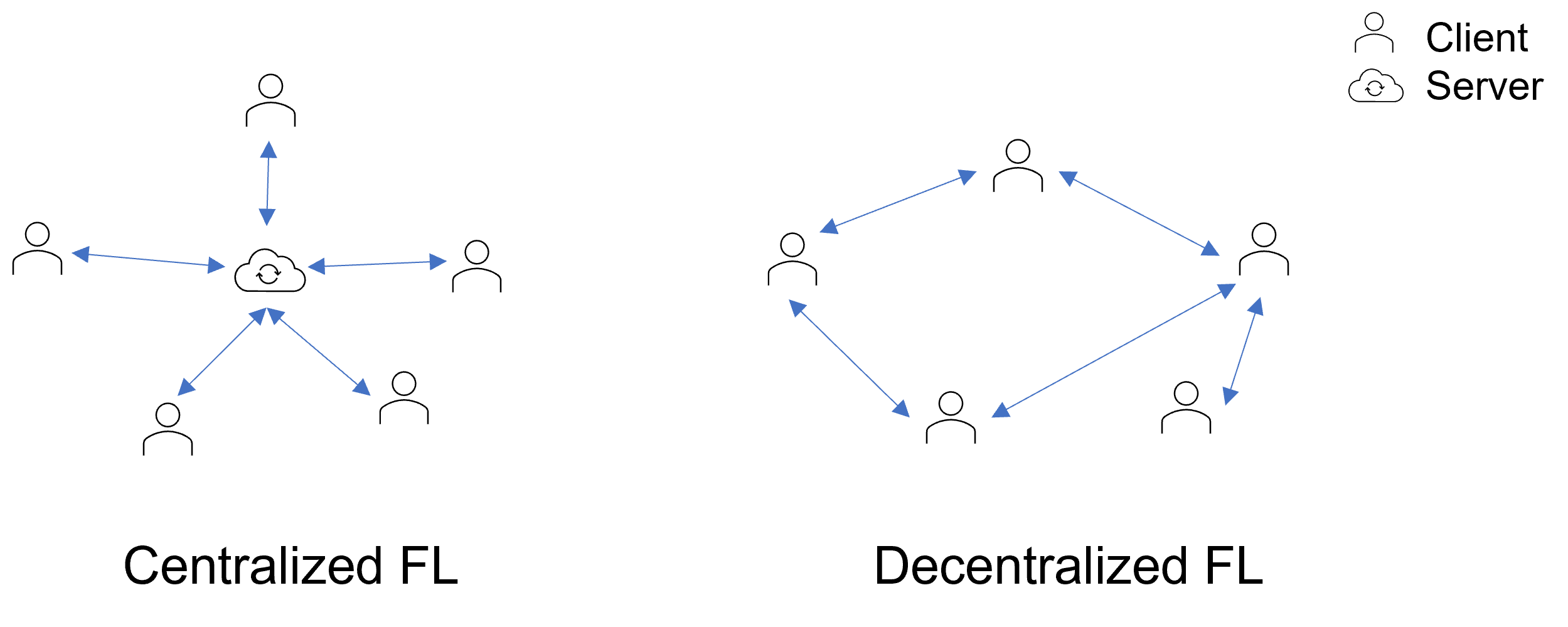}
    \caption{Communication topologies in FL:  centralized star configuration (left) versus decentralized peer-to-peer (right).} 
    \label{fig:topo}
\end{figure}

The underlying communication structure of FL can be modelled as an undirected graph  $G = (\mathcal{V}, \mathcal{E})$, where vertices $\mathcal{V} = \{1,\dots,n\}$ represent the nodes/agents/participants and  $\mathcal{E}=\{(i,j): i, j\in \mathcal{V}\}$ is the set of undirected edges in the graph representing the communication links, and let  $m=|\mathcal{E}|$ denote the total number of edges. 
Each node is only allowed to interact with its neighbouring nodes directly. Let $\mathcal{N}_i$ denote the set of neighbouring nodes of node $i$, i.e., $\mathcal{N}_i=\{j:(i,j)\in \mathcal{E}\}$ and $d_i = |{\mathcal N}_i|$.  Note that centralized FL is a special case of decentralized ones with a star topology $m=n-1$, i.e., one central server connects to all $n-1$ clients, as illustrated in  Figure~\ref{fig:topo}. .%Furthermore, we use the following notational conventions. A single subscript as in $x_i$ indicates that the variable is related to node $i$. A variable $x_{i|j}$ is related to edge $(i,j)$ but held by node $i$; it relates to the directed edge $i\to j$.  

\subsection{Problem formulation}
Each participant $i$ possesses a private dataset $\mathcal{D}_i = \{(\mathbf{x}_{ik}, \ell_{ik})\}_{k=1}^{n_i}$ where feature vectors $\mathbf{x}_{ik} \in \mathbb{R}^v$ correspond to labels $\ell_{ik} \in \mathbb{R}$. The local objective function $f_i(\mathbf{w}_i) = \frac{1}{n_i}\sum_{k=1}^{n_i} \phi(\mathbf{w}_i; \mathbf{x}_{ik}, \ell_{ik})$ measures model error, with $\mathbf{w}_i \in \mathbb{R}^u$ as learnable parameters and $\phi(\cdot)$ as an application-specific loss function. The global FL goal minimizes the collective objective $\frac{1}{n}\sum_{i=1}^n f_i(\mathbf{w}_i)$ while preserving data locality.

\subsection{Adversary model}
In scenarios involving multiple participants, two primary adversary models are typically considered \cite{damgaard2012multiparty}.

1) Passive adversary, often termed "honest-but-curious" adversary: it assumes that certain participants have been corrupted; although they follow the protocol correctly, they attempt to glean private information from any data they receive (e.g., via privacy attacks described in Section~\ref{subsec.passive}). When multiple participants are passively corrupt and collude, the risk of privacy breaches increases significantly.

2) Active adversary (often called "Byzantine adversary" in the context of FL): it poses a more severe threat. Actively corrupted participants can deviate from the prescribed protocol in arbitrary ways, such as poisoning local model updates, broadcasting falsified model parameters, or launching attacks aimed at disrupting or manipulating the global model (see Section~\ref{sec.active}).   %As a consequence, dealing with the active adversary is more challenging than the above passive adversary. That is, if an algorithm is robust to an active adversary implies that it is also robust to the passive adversary. The reverse is, however, not true.

\subsection{Security model} 
The security model defines the assumed computational capabilities of potential adversaries. Two main security models are commonly used: 
\begin{enumerate}
\item Computational security. The adversary is assumed to have bounded computational power and thus cannot feasibly decrypt protected data (i.e., not in polynomial time).
\item Information-theoretic security. The adversary is assumed to be computationally unbounded but lacks sufficient information to infer the secret.
\end{enumerate}
% \jane{add back later on if necessary} In this paper, we focus on information-theoretic security because it provides stronger protection and is generally more efficient in terms of both communication and computation \cite{lagendijk2013encrypted}.

\subsection{Performance evaluation}
The design of FL algorithms involves addressing several key aspects:\\
\noindent\textbf{Utility}: Ensuring that the model achieves high predictive performance, such as high test accuracy on unseen data.\\
\noindent\textbf{Privacy}: Safeguarding sensitive data of honest participants against the passive adversary (e.g., honest-but-curious participants).\\
\noindent\textbf{Robustness}: Ensuring system reliability and performance in the presence of active adversary (i.e., maliciously corrupt participants).\\
\noindent\textbf{Efficiency}: Minimizing the communication and computation overhead to enable practical deployment of FL systems.

%\subsubsection{Computational complexity}: Reducing the computational burden on participants to enable practical deployment of FL systems.

\section{Centralized FL}\label{sec.cfl}
In a typical CFL setting, a central server orchestrates the training process among multiple clients. We classify existing CFL protocols into two categories: protocols that separate local training and aggregation, and protocols that pose the entire framework as a joint optimization problem.

\subsection{Separate aggregation based protocols}\label{ssec.avgCFL}
A standard example of this category, often termed \emph{FedAvg}~\cite{mcmahan2017communication}, works as follows: 
\begin{enumerate}
    \item Initialization: at iteration  $t=0$, the server randomly initializes the global model weights \(\mathbf{w}^{(0)}\) and distributes them to each node. 
    \item Local model training: at each iteration $t$, each node \(i\) updates its local model \(\mathbf{w}_i^{(t)}\) based on \(\mathbf{w}^{(t-1)}\) using its local data \(\mathbf{x}_i\). These local updates are then sent to the server.
    \item Model aggregation: after collecting these local models, the server performs aggregation to update the global model. The aggregation is often done by weighted averaging and typically the weights are decided by the size of the local dataset $n_i$, i.e.,
\begin{align}\label{eq.w_ave}
   \mathbf  w^{(t)}=\tsum_{i=1}^{n}\frac{n_i}{\tsum_{k=1}^{n}n_k}\mathbf w_i^{(t)},
\end{align}
\end{enumerate}
\noindent
The last two steps  repeat until the global model converges or a stopping criterion is met.

Another widely used protocol, called \emph{FedSGD}~\cite{mcmahan2017communication}, assumes full-batch training on the client side. It closely resembles FedAvg in its overall structure, but differs in that clients share gradients $\mathbf{g}_i^{(t)}=\nabla f_i(\v w_i)$ instead of sharing the weights, i.e. 
\begin{align}\label{eq.g_ave}
   \mathbf  w^{(t+1)}=\mathbf w^{(t)}- \mu\tsum_{i=1}^{n}\frac{n_i}{\tsum_{k=1}^{n}n_k} \mathbf{g}_i^{(t)},
\end{align}
where $\mu$ is a stepsize controlling the convergence rate. 

These methods rely on alternating rounds of local model training and server-side aggregation, forming the backbone of many FL protocols. Building on them, many extensions have been proposed to enhance either local training or the aggregation procedure. For instance, Per-FedAvg~\cite{fallah2020personalized}, MOON~\cite{li2021model}, FedProx~\cite{li2020federated}, FedNova~\cite{wang2020tackling} are proposed to address challenges in local training step such as suboptimal performance, slow convergence, or model drift caused by data heterogeneity.  Alternatively, other modifications focus on improving the robustness of the aggregation step, which will be further explained in Section~\ref{sec.robust}.

\subsection{Joint optimization based protocols}
Instead of having separated local training and model aggregation, \emph{joint optimization-based protocols} treat the entire FL process as a single optimization problem. By unifying the local update step with the global objective, these methods can leverage classical distributed optimization techniques such as ADMM~\cite{boyd2011distributed,giselsson2016linear,shi2014linear} and PDMM~\cite{zhang2018distributed,sherson2018derivation,heusdens2024distributed,heusdens2024distributed2}. ADMM and PDMM have been widely adopted in various applications such as acoustic signal processing \cite{sherson2016distributed,koutrouvelis2018low}, privacy-preserving data aggregation~\cite{Jane2020ICASSP,Jane2020LS,li2020privacy,huang2015differentially,han2016differentially,li2019privacyA,nozari2018differentially,zhang2016dynamic,zhang2018recycled,zhang2018improving,xiong2020privacy,jane2022gmm,jordan2024,li2022communication,li2023adaptive,yu2024privacy} and also in FL~\cite{niwa2020edge,yu2024provable,pathak2020fedsplit,karimireddy2020scaffold,zhang2022revisiting,li2024topology}.  

\paragraph{Star-graph formulation:} We can model the CFL network as a special star graph: $\mathcal{G}=(\mathcal{V},\mathcal{E})$, with $\mathcal{V}={\{0,1,2,...,n}\}$ representing the node set and $\mathcal{E}\subseteq \mathcal{V}\times \mathcal{V}$ representing the edge set.  For notation simplicity, denote node $0$ serves as the central server.  In this case, for any client $i \in \mathcal{V}\setminus\{0\}$, its neighborhood $\mathcal{N}_i$ only contains the server node $0$.  Thus we can formulate the problem as 
\begin{align}
\begin{array}{ll}
{\min \limits_{\big\{\mathbf w_i \,:\, i\in \mathcal V\big\}}} &\tsum\limits_{i \in \mathcal{V}\backslash\{0\}} f_i(\mathbf w_i), \\
\text{subject to} &\mathbf{w}_0 = \mathbf{w}_i~~ \; i=1,\ldots, n, \rule[4mm]{0mm}{0mm}
\end{array}
\end{align}

Popular algorithms of this type include  FedSplit~\cite{pathak2020fedsplit}, SCAFFOLD~\cite{karimireddy2020scaffold}, and GPDMM/AGPDMM~\cite{zhang2022revisiting}. In particular, \textsc{GPDMM} can serve as a general representative of this class.    Furthermore,  SCAFFOLD and GPDMM can both reduce to vanilla gradient descent \cite{ruder2016overview} under appropriate parameter configurations.

Below, we describe the main procedure of FedSplit and GPDMM: 
\begin{enumerate}
    \item Initialization: at iteration  $t=0$, each node initializes the weights $\mathbf w_i^{(0)}=\mathbf w_0^{(0)}$.
    \item Local model update: at each iteration $t$, each user $i$ first updates its local model $\mathbf w_i^{(t)}$ and variable $\mathbf z_{i|0}^{(t)}$ with the following \cref{equ:client_update}. Then each user sends $\mathbf z_{i|0}^{(t)}$ to the server.
    \begin{align}
& \textrm{users}\left\{ \begin{array}{l}
\mathbf{w}_i^{(t+1)} =\arg\min_{\mathbf{w}_i}\Big[f_i(\mathbf{w}_i) +\frac{1}{2\gamma}\|\mathbf{w}_i-\mathbf{z}_{0|i}^{(t)} \|^2 \Big]  \\
\mathbf{z}_{i|0}^{(t+1)} = 2 \mathbf{w}_i^{(t+1)} 
-\mathbf{z}_{0|i}^{(t)} \end{array}\right. \label{equ:client_update} 
\end{align}
where $\v z$ is the auxiliary variable and regularization parameter $\gamma>0$.
    \item Model aggregation: the server collects these $\v z_{i|0}^{(t)}$ and performs aggregation to update the global model $\mathbf w_0^{(t)}$. The aggregation is done by following \cref{equ:server_update}. Then the server calculates and sends $\mathbf z_{0|i}^{(t)}$ to each user.
    \begin{align}
& \textrm{server} \left\{ \begin{array}{l}
\mathbf{w}_0^{(t+1)}=\frac{1}{n}\sum_{i=1}^n \mathbf{z}_{i|0}^{(t+1)}  \\
\mathbf{z}_{0|i}^{(t+1)} = 2\mathbf{w}_0^{(t+1)} 
-\mathbf{z}_{i|0}^{(t+1)}  \end{array}\right.  \label{equ:server_update}
\end{align}
\end{enumerate}
\noindent
The last two steps are repeated until the global model converges or until a predetermined stopping criterion is reached.

\paragraph{Inexact Updates:} In practice, computing the exact solution in Equation \eqref{equ:client_update} can be computationally demanding or prohibitively expensive. Consequently, two distinct approximation techniques are commonly employed to address this challenge.

\textbf{$K$ gradient descent:}
 A commonly employed technique is to approximate each local subproblem using $K$ gradient descent steps:
\begin{align}\label{equ:inexact_iterate}
    \mathbf{w}_i^{(t,k+1)} =\mathbf{w}_i^{(t,k)}-\gamma \nabla h_i^{(t)}(\mathbf{w}_i^{(t,k)}), 0\leq k <K,
\end{align}
where the augmented local objective $h_i^{(t)}(\mathbf{w}_i)$ is defined as $h_i^{(t)}(\mathbf{w}_i)=f_i(\mathbf{w}_i) +\frac{1}{2\gamma}\|\mathbf{w}_i-\mathbf{z}_{0|i}^{(t)} \|^2$.

Notably,
Zhang {\it et al.}~\cite{zhang2022revisiting} demonstrate that PDMM can reduce to FedSplit in CFL; however, FedSplit suffers from poor performance due to improper parameter initialization. Specifically, FedSplit sets  $\mathbf{w}_i^{(t,k=0)}=\mathbf{z}_{0\mid i}^{(t)}$ at each iteration. In contrast, GPDMM achieves improved stability by reusing the final updates from the previous iteration: $\mathbf{w}_i^{(t,k=0)}=\mathbf{w}_i^{(t-1,k=K)}$.

\textbf{Quadratic approximation:}
An alternative strategy is to use quadratic approximation~\cite{o2018function}, which serves as a viable and computationally efficient method for approximating $\mathbf w_i^{(t)}$.

The function $f(\mathbf{w})$ can be approximated using a quadratical surrogate:
\begin{equation}\label{eq.qa}
f(\mathbf{w}) \approx f\left(\mathbf{w}^{(t)}\right)+\nabla f\left(\mathbf{w}^{(t)}\right)^\intercal\left(\mathbf{w}-\mathbf{w}^{(t)}\right)+\frac{\mu}{2}\left\|\mathbf{w}-\mathbf{w}^{(t)}\right\|^2,
\end{equation}
where \(\mu > 0\) is chosen large enough to keep the surrogate strongly convex. Under this approximation, the update on node \(i\) simplifies to
\begin{align}
        \nonumber \mathbf w_i^{(t+1)}&=\frac{\gamma\mu \mathbf w_i^{(t)}-\gamma\nabla f_i(\mathbf w_i)+\mathbf z_{i \mid 0}^{(t)} }{1+\gamma\mu}. 
    \end{align}
In fact, as will be discussed in Section \ref{ssec.equ_cfl_dfl}, the formulation in CFL represents a specific instance of the more general DFL framework \eqref{eq.qa_update}. Consequently, the convergence guarantee provided in Section \ref{ssec.optDFL} is also applicable to this case.

\section{Decentralized FL}\label{sec.dfl}
In this decentralized setup, there is no central server to coordinate model updates; instead, each node iteratively exchanges information with its neighbors to learn a global model.
\subsection{Separate aggregation-based approaches}\label{ssec.avgDFL}
A straightforward way to aggregate local models in DFL is through distributed average consensus methods. Rather than sending updates to a central server, each node combines the model parameters received from its neighbors. Examples of consensus algorithms include randomized gossip~\cite{dimakis2010gossip} and linear iterations~\cite{olshevsky2009convergence}. 

Concretely, many DFL protocols closely resemble FedAvg except for how the model aggregation step is performed. For instance, the approaches proposed in~\cite{lian2017can,koloskova2020unified,sun2022decentralized} use gossip averaging to implement the model aggregation, i.e., 
\begin{align}\label{eq.dfl_sa}\mathbf w_i^{(t+1)}=\tsum_{j\in \mathcal{N}_i\cup i}A_{ij}\mathbf w_j^{(t)},
\end{align}
where $A_{ij}$ is the $i$-th and $j$-th element of weight matrix $\v A$. The matrix \(\v A\)  is typically chosen to be: 1) Nonnegative  $A_{ij}\in [0,1]$, which $0$ means disconnected; 2) symmetric $\v A^{\top}=\v A$; 3) doubly stochastic $\v A \mathbf{1}=\mathbf{1}$ and  $\mathbf{1}^{\top}\v A=\mathbf{1}^{\top}$. Under these conditions, iterative gossip-based updates allow each node to converge toward a consensus model over iterations.

\subsection{Joint optimization-based approaches} \label{ssec.optDFL}
Although averaging-based strategies are intuitive, they can suffer from slow convergence or local minima issues, particularly in the presence of heterogeneous data~\cite{niwa2020edge}. An alternative is to formulate DFL as a single constrained optimization problem over the entire network:
\begin{align} \label{eq.pmFor}
\begin{array}{ll}
{\min \limits_{\big\{\mathbf w_i \,:\, i\in \mathcal V\big\}}} &\tsum\limits_{i \in \mathcal{V}} f_i(\mathbf w_i), \\
\text{subject to} &\forall (i,j) \in {\mathcal E}:{{B}_{i\mid j}}{{\mathbf w}_i} + {{B}_{j\mid i}}{{\mathbf w}_j} = \v 0, \rule[4mm]{0mm}{0mm}
\end{array}
\end{align}
where the linear constraints \(B_{i\mid j}\,\mathbf w_i + B_{j\mid i}\,\mathbf w_j = \mathbf{0}\) ensure that, at convergence, all nodes share a common model (consensus). For simplicity, one can set \(B_{i\mid j} = - B_{j\mid i} = \pm 1\) or a suitable identity-matrix variation.

\paragraph{ ADMM/PDMM Protocols:}
The common unicast ADMM/PDMM protocol~\cite{niwa2020edge} using ADMM~\cite{giselsson2016linear} or PDMM~\cite{zhang2018distributed,sherson2018derivation} work as follows: 
\begin{enumerate}
    \item Initialization: at iteration  $t=0$, each node randomly initializes the weights $\mathbf w_i^{(0)}$.
    \item Local model training: at each iteration $t$, each user $i$ first updates its local model $\mathbf w_i^{(t)}$ and variable $\mathbf z_{j|i}^{(t)}$ with the following \cref{equ:dfl_client_update}. Then each user sends $\mathbf z_{j|i}^{(t)}$ to their neighbors $j \in \mathcal{N}_i$.
    \begin{align}\label{equ:dfl_client_update}
& \textrm{users}\left\{ \begin{array}{l}
\mathbf{w}_i^{(t+1)} =\arg\min_{\mathbf{w}_i}\Big[f_i(\mathbf{w}_i) +\sum_{j \in \mathcal{N}_i}\mathbf{z}_{i \mid j}^{(t)\top} B_{i\mid j} \mathbf{w}_i+\frac{cd_i}{2}\left\|\mathbf{w}_i\right\|^2 \Big]  \\
\mathbf{z}_{j|i}^{(t+1)} = (1-\theta)\mathbf{z}_{j|i}^{(t)}+\theta(\mathbf{z}_{i|j}^{(t)}+2cB_{i\mid j} \mathbf{w}_i^{(t+1)})\end{array}\right. 
\end{align}
where $\theta\in (0,1]$, with $\theta=1$ corresponding to PDMM and $\theta=\frac{1}{2}$ corresponding to ADMM.
\end{enumerate}
\noindent
The last step is repeated until the global model converges or until a predetermined stopping criterion is reached.

\textbf{Quadratic approximation:}
In DFL, \eqref{eq.qa} can be simplified to
\begin{align}\label{eq.qa_update}
        \mathbf w_i^{(t+1)}&=\frac{\mu \mathbf w_i^{(t)}-\nabla f_i(\mathbf w_i^{(t)})-\sum_{j \in \mathcal{N}_i}B_{i\mid j}\mathbf z_{i \mid j}^{(t)} }{\mu +cd_i}. 
    \end{align}
Given the following theorem, the above is guaranteed to converge~\cite{o2018function}.
\begin{theorem}
    Assume $f(\mathbf{w})$ is $m$-strongly convex and $\beta$-smooth, then as long as $$\mu>\frac{\beta^2}{2m},$$ we have 
    $$\lim_{t\rightarrow\infty}\mathbf{w}^{(t)}=\mathbf{w}^*.$$   
\end{theorem}

\subsection{On equivalence between CFL and DFL} \label{ssec.equ_cfl_dfl}
A key insight from our formulations is that, for both separate-aggregation and joint-optimization frameworks,  CFL can be viewed as a special case of DFL by simply restricting the topology to a star-shaped network. 

\noindent
\textbf{Separate aggregation based CFL and DFL:}
When a central server is available, each client communicates with the server directly. In DFL terms, this corresponds to a star network, where the weight-update rule in \cref{eq.dfl_sa} can be rewritten as
\begin{align}\mathbf w_i^{(t+1)}=\tsum_{j\in \mathcal{V}}A_{ij}\mathbf w_j^{(t)},
\end{align}
since the server has access to all $\mathbf w_j^{(t)}$ for every $j \in \mathcal{V}$, it can aggregate these updates and redistribute them to the clients. If we set \(A_{ij} = \frac{n_j}{\tsum_{k=1}^{n}n_k}\), then the update precisely reduces to the weighted average \(\mathbf w^{(t)}\) in \cref{eq.w_ave}, recovering the standard FedAvg procedure.

\noindent
\textbf{Joint-optimization based CFL and DFL:}
Again let node \(0\) denote the server. The primal-dual approach in CFL, i.e., \cref{equ:client_update} and \cref{equ:server_update} are actually the specific case of \cref{equ:dfl_client_update}, when the edge set $\mathcal{E}$ in the graph is $E = \{(i, 0)\}_{i=1}^n$ and the server
function $f_0(\mathbf w_0) = 0$. By choosing $B_{i\mid 0}=-1, B_{0\mid i}=1, \gamma=\frac{1}{c}, \mathbf{z}_{0\mid i}=\frac{1}{c}\mathbf{z}_{i\mid j}$ and $ \mathbf{z}_{i\mid 0}=-\frac{1}{c}\mathbf{z}_{j\mid i}$, the updates in the CFL formulation match those in \cref{equ:dfl_client_update} exactly, confirming the equivalence.

In Table~\ref{table:fl_form} we summarize existing CFL and DFL algorithms based on the above formulations. 

\begin{table*}[h!]
\centering
\caption{An overview of CFL and DFL algorithms: separate aggregation vs. joint optimization}
\label{table:fl_form}
\begin{tabular}{p{1cm}|p{6.5cm}|p{6.5cm}}
\bottomrule
\textbf{}& \textbf{Separate Aggregation} & \textbf{Joint Optimization}\\
\hline
\textbf{CFL}& FedAvg/FedSGD~\cite{mcmahan2017communication}, Per-Fed~\cite{fallah2020personalized}, FedProx~\cite{li2020federated}, FedNova~\cite{wang2020tackling}, MOON~\cite{li2021model} & FedSplit~\cite{pathak2020fedsplit}, GPDMM/AGPDMM~\cite{zhang2022revisiting}, SCAFFOLD~\cite{karimireddy2020scaffold}\\
\hline
\textbf{DFL}& D-PSGD~\cite{lian2017can}, Braintorrent~\cite{roy2019braintorrent}, Combo~\cite{hu2019decentralized}, $D^2$~\cite{tang2018d}, GossipSGD~\cite{koloskova2020unified}, DFedAvgM~\cite{sun2022decentralized} & ADMM/PDMM~\cite{niwa2020edge,yu2024provable}\\
\toprule
\end{tabular}
\end{table*}

\section{FL against Passive Adversary: Attacks and Defenses}\label{sec.passive}
Though FL is designed to safeguard user privacy by eliminating the need of sharing raw training data, yet adversaries can still exploit shared model information to breach privacy. Especially, when consider the passive adversary that the corrupt participants can using their receives updates to conduct privacy breaches. In what follows we will summarize existing privacy attacks and the corresponding defenses.
\subsection{Empirical privacy attacks} \label{subsec.passive}
We review three main forms of privacy attacks: membership inference~\cite{melis2019exploiting,shokri2017membership}, property inference~\cite{xu2020subject,melis2019exploiting,wainakh2021user} and input reconstruction attack~\cite{he2019model,wang2019beyond,yang2019neural}.

\subsubsection{Membership inference attack (MIA)}
The goal of MIA  is to determine   whether a specific data sample was used during model training. There are many MIAs have been proposed in the last decade \cite{shokri2017membership,yeom2018privacy,salem2018ml,song2021systematic,melis2019exploiting,nasr2019comprehensive,hu2021membership,hu2023loss}.  A canonical strategy in designing MIA is to employ construct the so-called  shadow models that reproduce the behavior of the target model, enabling the attacker to infer membership by comparing outputs on various inputs.

In FL, recent work~\cite{li2022effective} has shown that gradient-based membership inference, which measures the cosine similarity between guessed and actual gradients, can surpass classic loss- or entropy-based approaches~\cite{yeom2018privacy,song2021systematic}. The main idea is to leverag the cosine similarity between model updates and instance-specific gradients for distinguishing training data from the others, which is formulated as:
\begin{align}
&M\left(\mathbf{x}^{\prime}, i\right)=\tsum_{l' \in \mathbb{R}} \mathbb{1}\left\{\operatorname{cosim}\left(\nabla f_i(\mathbf w_i,(\mathbf x', \mathbf\ell')), \nabla f_i(\mathbf w_i,(\mathbf x_i,\mathbf\ell_i))\right) \geq \gamma \right\},
\end{align}
where $\gamma$ is the threshold. A variant known as \emph{subject MIA}~\cite{suri2022subject,li2024subject} determines whether an individual user’s data contributed to training by observing changes in the global loss.
Subject MIA operates effectively in a black-box setting, leveraging the statistical distribution of sample features and the loss function to infer the presence of an individual in the training dataset.

\subsubsection{Property inference attack (PIA)}
Property inference attacks aim to reveal hidden attributes of the training set. It can be broadly classified into two categories: passive~\cite{melis2019exploiting, wainakh2021user,kerkouche2023client} and active~\cite{melis2019exploiting,xu2020subject,kim2023exploring, wang2022poisoning} attacks. In FL framework, model updates or gradients are often utilized to infer specific properties or attributes of the data~\cite{xu2020subject,melis2019exploiting,wainakh2021user,wang2022poisoning,kerkouche2023client}. These attributes can vary widely, ranging from uncorrelated information, identifying whether a photo subject is wearing eyeglasses in a model trained for gender or race~\cite{melis2019exploiting}, to extracting labeling information about the data itself~\cite{wainakh2021user}.

\subsubsection{Input reconstruction attack (IRA)} 
Input reconstruction goes beyond membership or property inference, striving to recover entire training samples. 
Based on the available information, there are broadly two ways to reconstruct the input training data.  

\noindent\textbf{Gradient inversion attack:} 
The most common and extensively studied method for reconstructing data in the FL framework is the gradient inversion attack \cite{zhu2019deep}. It leverages gradient information or its approximations, which can be readily obtained from transmissions over the communication links or the corrupt participants, for reconstructing the inputs.

The core mechanism behind gradient inversion attacks is an iterative process that refines a guessed input to match the gradients observed during training. Formally, for each client’s local dataset \((\mathbf{x}_i,\mathbf{\ell}_i)\), the adversary attempts to recover the true data by solving~\cite{zhu2019deep}:
\begin{align}\label{eq.traInv}
(\mathbf x_i^{\prime *}, \mathbf \ell_i^{\prime *}) =\underset{\mathbf x_i^{\prime}, \mathbf \ell_i^{\prime}}{\arg \min }\big\|\nabla f_i(\mathbf w_i,(\mathbf x'_i,\mathbf \ell'_i)) - \nabla f_i(\mathbf w_i,(\mathbf x_i,\mathbf \ell_i))\big\|^2,
\end{align} 
with many subsequent variants proposed~\cite{zhao2020idlg,geiping2020inverting,yin2021see,boenisch2021curious,yang2022using,li2024perfect}. 

Although FL typically shares model weights rather than raw gradients (see Sections \ref{ssec.avgCFL} and \ref{ssec.avgDFL}), adversaries can still approximate gradients using the differences in the model parameters, \(\mathbf{w}_i^{(t+1)} - \mathbf{w}_i^{(t)}\). In fact, works such as~\cite{xu2022agic,geng2023improved,zhao2022deep} demonstrate how these approximate gradients can reveal sensitive data. Moreover, Liang \emph{et al.} ~\cite{10209197} proposes a scenario where an intermediary attacker intercepts real gradients and injects spurious queries to further refine its estimates, thereby obviating the need for prior knowledge of model weights.

\noindent\textbf{Output-based inversion attack:} 
As the name suggests, the main idea of output-based inversion attack is to reconstruct training data using only the model and its outputs (e.g., probability distributions). Even in FL settings, adversaries can observe the updates and reverse-engineer the inputs by analyzing the distributional patterns. This is typically achieved by training a secondary model designed to act as an inverse of the original model~\cite{yang2019neural, yang2019diversity, zhang2020secret, chen2021knowledge, zhang2023analysis}.

We now proceed to defense mechanisms that protect FL from the privacy attacks discussed earlier.  These methods broadly fall into two categories: those offering \emph{provable privacy guarantees} and those providing \emph{empirical defenses} without formal security proofs. Table~\ref{tab:approaches_comparison} summarizes representative methods in each category. 

\subsection{Provable defense approaches}

\subsubsection{Differential Privacy (DP)}
DP offers probabilistic guarantees against privacy breaches by injecting calibrated noise into model updates or outputs. Under classical DP assumptions, all but one node in the network may be considered compromised~\cite{dwork2006,dwork2009differential}.  To formalize this, let $\v s^{-i} \in \mathbb{R}^{n-1}$ represent an adjacent vector of $\v{s}$, obtained by excluding the private data $s_i$ from $\v s$. Denote the range of $s_i$ as $\Omega_i$. Let $\hat{F}$ be a randomized algorithm designed to preserve the privacy of its input, and let $\mathcal{Y}$ denote its output space. 
% Given a privacy parameter $\epsilon \geq 0$, the algorithm $\hat{F}$ satisfies $\epsilon$-DP if, for any pair of adjacent vectors $\v{s}$ and $\v{s}^{-i}$, and for all subsets $\mathcal{Y}_s \subseteq \mathcal{Y}$, the following condition holds:
% \begin{align}
%     \forall s_i \in \Omega_i: \quad 
%     \frac{P(\hat{F}(\v{s}) \in \mathcal{Y}_s)}{P(\hat{F}(\v{s}^{-i}) \in \mathcal{Y}_s)} \leq e^{\epsilon}. \label{eq.oriDP}
% \end{align}
Let \(\mathbf{s}^{-i}\in \mathbb{R}^{n-1}\) represent a vector adjacent to \(\mathbf{s}\) but missing the private data \(s_i\), and let \(\Omega_i\) be the range of \(s_i\). A randomized algorithm \(\hat{F}\) providing \(\epsilon\)-DP must satisfy, for any \(\mathbf{s}\) and its adjacent \(\mathbf{s}^{-i}\) and for all subsets \(\mathcal{Y}_s \subseteq \mathcal{Y}\),
\begin{align}
    \forall s_i \in \Omega_i:\quad 
    \frac{P\bigl(\hat{F}(\mathbf{s}) \in \mathcal{Y}_s\bigr)}
         {P\bigl(\hat{F}(\mathbf{s}^{-i}) \in \mathcal{Y}_s\bigr)} 
    \;\leq\; e^{\epsilon}.\label{eq.oriDP}
\end{align}
This criterion ensures that the algorithm’s outputs are indistinguishable whether or not \(s_i\) is included. Local DP (LDP) applies noise directly to each client’s data~\cite{duchi2013local}, making it suitable for untrusted servers, while Global DP (GDP) adds noise at the server level~\cite{abadi2016deep}, requiring a trusted central authority. In practice, tuning DP’s privacy–utility trade-off can be challenging, especially for non-convex models such as deep neural networks, and often relies on heuristic noise schedules~\cite{niwa2020edge} or adaptive clipping methods~\cite{yu2024provable}.

\subsubsection{Secure multiparty computation (SMPC)}  
SMPC allows a set of parties to jointly compute a function over their inputs while keeping those inputs private from each other\cite{damgaard2012multiparty}. SMPC techniques are frequently integrated with differential privacy mechanisms to advance the development of secure FL frameworks.
In CFL, Gong {\it et al.}~\cite{gong2020privacy} introduce a privacy-preserving multi-party framework which leverages the integration of differential privacy and homomorphic encryption to mitigate the risk of privacy breaches. This approach eliminates the necessity for a server who must be universally trusted by all participants involved. Zhou {\it et al.}~\cite{zhou2023decentralized} propose a communication mechanism based on secret sharing to enable encrypted decentralized federated learning. It introduces a secure stochastic gradient descent scheme combined with an autoencoder and a Gaussian mechanism to anonymize local model updates, facilitating secure communication among a limited number of neighboring clients. Tran {\it et al.}~\cite{tran2021efficient} also integrate randomization techniques with an efficient secret sharing protocol to mitigate potential security threats. Their approach emphasizes safeguarding local models against honest-but-curious adversaries, even in scenarios where up to $n-2$ out of $n$ parties may collude.

\subsubsection{Subspace Perturbation}
Subspace perturbation \cite{Jane2020ICASSP,Jane2020LS,li2020privacy,jordan2024,li2022communication,li2023adaptive} takes a different route by masking gradients in directions orthogonal to the primary optimization trajectory. 
Its main idea is to exploit the structure of distributed optimization algorithms to confine noise to the so-called non-convergent subspaces. By perturbing updates orthogonal to the optimization trajectory, these methods mask sensitive gradients without degrading accuracy. Unlike DP, it requires no explicit noise calibration and operates efficiently over incomplete networks. Recent work demonstrates its effectiveness in DFL~\cite{yu2024provable} for protecting privacy.

\subsection{Empirical defense approaches}
Empirical defenses, while lacking formal guarantees, offer practical mitigation against specific attacks. 
Given that the shared information such as models or weights carries sensitive information, a natural defense is thus to compress the data using techniques such as quantization etc.

\subsubsection{Gradient sparsification}
Several studies~\cite{jeon2021gradient, huang2021evaluating} have demonstrated that commonly used gradient sparsification techniques~\cite{aji2017sparse, bernstein2018signsgd} not only effectively reduce communication overhead but also provide a certain level of defense against various privacy attacks. These sparsification methods typically operate by transmitting only a subset of the gradient values, such as selecting the top K highest-magnitude elements while setting the remaining values to zero. This selective transmission reduces the amount of data exchanged during training, thereby improving efficiency and mitigating potential privacy risks.

However, recent research~\cite{melis2019exploiting, fu2022label, huang2021evaluating, wu2023learning} has examined the effectiveness of gradient sparsification from the perspective of novel attack strategies and has identified potential vulnerabilities. These studies suggest that while gradient sparsification may offer some level of privacy protection, it does not always provide robust defense against adversarial attacks. 

\subsubsection{Gradient quantization}
Quantization is also a technique to reduce the communication cost by reduces the number of bits used to represent each gradient value. One thing to note is that applying quantization can also be viewed as an information theoretical approach as it will introduce distortion or noise to the transmitted messages. This can be analog to differential privacy noise addition as long as the introduced noise by quantization is independent of the message, which can be guaranteed by dithering.

Both~\cite{lang2023joint} and \cite{kang2024effect} highlight that gradient quantization inherently aligns with DP or its variants in the context of CFL. Similarly, Li {\it et al.}~\cite{Jane2022twoforone, li2023adaptive} provide a theoretical justification for this phenomenon in the setting of DFL. These findings actually suggest a fundamental connection between gradient quantization techniques and informational theoretical mechanisms in FL frameworks, reinforcing their role in enhancing both communication efficiency and data privacy.

In addition to data sparsification techniques, another practical defense is \emph{data augmentation}. 
\subsubsection{Data augmentation}
It has been demonstrated that data augmentation techniques can mitigate input reconstruction attacks~\cite{shin2023empirical}. In particular, Shin {\it et al.}~\cite{shin2023empirical} find that certain augmentation schemes not only resist input reconstruction but also outperform DP in model accuracy, highlighting a compelling trade-off between privacy protection and predictive performance.

\begin{table}[h!]
\centering
\scriptsize
\begin{tabular}{l|l|c|c|c|c|cc|}
\bottomrule
\textbf{Approaches} & \textbf{\makecell[c]{Assumed \\Topology}} & \textbf{\makecell[c]{General \\Applicable}} & \textbf{\makecell[c]{Provable\\ Privacy}} & \multicolumn{3}{c}{\textbf{Empirical Privacy Attacks}} \\
\cline{5-7}
 & & & & \textbf{\makecell[c]{Membership \\Inference}} & \textbf{\makecell[c]{Property \\Inference}} & \textbf{\makecell[c]{Input \\Reconstruction}} \\
\hline
Differential privacy & CFL \cite{gong2020privacy,wei2020federated,hu2020personalized,truex2020ldp,triastcyn2019federated} / DFL \cite{cyffers2022privacy} & Yes & Yes & \cite{truex2020ldp,naseri2020local} & \cite{ren2022grnn,feng2022user} & \cite{ren2022grnn,wu2023learning,huang2021evaluating} \\
Subspace perturbation & DFL \cite{yu2024provable} &No & Yes & \cite{yu2024provable} & - & \cite{yu2024provable,li2024privacy} \\
SMPC & CFL \cite{gong2020privacy} / DFL \cite{tran2021efficient,zhou2023decentralized} & Yes & Yes & \cite{gehlhar2023safefl,kaminaga2023mpcfl,liu2021privacy} & \cite{gehlhar2023safefl,kaminaga2023mpcfl,liu2021privacy} & \cite{kaminaga2023mpcfl,liu2021privacy,zhang2022augmented,geng2024privacy} \\
Data augmentation & CFL \cite{shin2020xor} & Yes & No & \cite{zhu2024evaluating} & - & \cite{shin2023empirical} \\
Gradient sparsification & CFL \cite{aji2017sparse,bernstein2018signsgd} & Yes & No & \cite{melis2019exploiting,zhu2024evaluating} & \cite{fu2022label} & \cite{zhu2019deep,jeon2021gradient,wu2023learning,huang2021evaluating} \\
Gradient quantization & CFL \cite{shlezinger2020uveqfed,tonellotto2021neural} / DFL \cite{li2023adaptive} & Yes & Yes (relaxed DP) & \cite{li2022effective,elkordy2022heterosag,kang2024effect,zhu2024evaluating} & - & \cite{ovi2023mixed,lang2023joint,elkordy2022heterosag} \\
\toprule
\end{tabular}
\caption{Comparison of Approaches in Privacy-Preserving Federated Learning.}
\label{tab:approaches_comparison}
\end{table}

\subsection{Gap between provable privacy guarantee and empirical privacy attacks}
A significant challenge in privacy research is reconciling provable privacy guarantees with empirical privacy attack results. For example, Recent work highlights that many MIAs produce misleading results that fail to comply with the concept of DP~\cite{li2024privacy}. Such inconsistencies arise from several factors, including simplified attack models, the real-world trade-offs involved in implementing DP, and the metrics used to measure privacy leakage~\cite{nasr2021adversary, jayaraman2019evaluating, li2024privacy, carlini2022membership}.   %Empirical attacks may exploit factors like overfitting and training dynamics, which are not always accounted for in theoretical frameworks \cite{zhang2021understanding}.  

DFL, however, may help narrow this gap by eliminating the need for a central server and exchanging partially obfuscated gradients directly among peers. This design inherently reduces the adversary’s visibility of sensitive data and can provide provable privacy advantages over CFL, as demonstrated in~\cite{yu2024provable}.  By reducing adversarial access to sensitive information and leveraging distributed optimization, DFL contributes to narrowing the gap between provable privacy guarantees and empirical attack resilience, aligning theoretical assurances more closely with practical outcomes~\cite{ji2024re}.

\subsection{The debate of privacy comparisons of CFL and DFL}
Conflicting perspectives exist regarding which paradigm—CFL or DFL—ultimately offers stronger privacy protections. Pasquini \emph{et al.}~\cite{security2023pasquini} contend that DFL is less private than CFL, based on experiments involving loss-based MIAs under separated aggregation. They attribute DFL’s heightened vulnerability to the inhomogeneity of decentralized topologies, where node-specific training can amplify privacy leakage.   In contrast, Yu {\it et al.}~\cite{yu2024provable} argue that DFL can surpass CFL from an information-theoretic viewpoint, particularly when joint optimization protocol is employed. They note that CFL’s reliance on global aggregations creates an avenue for direct gradient exposure. Similarly, Ji \emph{et al.}~\cite{ji2024re} investigate this discrepancy from a mutual-information-based framework, illustrating how secure aggregation and other design choices can affect privacy outcomes in both CFL and DFL. Taken together, these conflicting results underscore the complexity of comparing privacy across federated paradigms and suggest that the choice between CFL and DFL may depend on specific system requirements, network configurations, and threat models.

\section{ FL against active adversary I: empirical attacks}\label{sec.active}
Active adversaries do more than passively observe FL updates; they deliberately manipulate data or model updates to disrupt training or embed hidden malicious behavior. Active attacks can be broadly classified into two categories: explicit attacks that degrade the global model’s performance or prevent its convergence, and backdoor attacks that inject hidden triggers into the model while maintaining high utility on normal inputs.

\subsection{Explicit attacks: compromising model utility}
In a CFL setup, the typical assumption is that the server is trusted and only clients might be malicious. Thus, most active attacks studied in the literature assume malicious clients (sometimes called Byzantine clients). These clients send incorrect or adversarial updates to the server in order to bias the aggregation. In contrast, the server is usually assumed to have a high security level and is harder to corrupt, meaning it does not directly inject malicious updates. We summarize several representative {explicit attack strategies below:

\textbf{A Little Is Enough (ALIE)~\cite{baruch2019little}:} This stealthy attack leverages the statistical properties of honest updates to remain undetected. Let $\mu_h$ and $\sigma_h$ denote the mean and standard deviation of the gradient vectors $\mathbf g_h$ from honest clients $h$. A Byzantine client $m$ draws its malicious gradient $\mathbf g_m$ by slightly shifting the honest mean:
\begin{align}
    \mathbf g_m = \mu_h - \epsilon\, \sigma_h,
\end{align}
where $\epsilon$ controls the attack intensity. By perturbing the gradients only modestly, ALIE shifts the global model in an adversary-chosen direction while evading many outlier detection schemes.

\textbf{Bit Flipping (BF) \cite{Rakin_2019_ICCV,karimireddy2020byzantine}:} In this straightforward attack, a malicious client simply flips the sign of its gradient update to oppose the direction of honest updates. Formally, $\mathbf g_m = -\,\mathbf g_h$. BF can significantly slow down or derail the training process if not mitigated. However, it is relatively easy to detect or counter by defenses that check for abnormal update magnitudes or directions (e.g., by clipping large updates or using robust aggregators).

\textbf{Mimic Attack ~\cite{zhang2019poisoning}:} Here, malicious clients attempt to appear benign by crafting their updates to statistically resemble honest updates. An attacker may run a "warm-up" period to learn the distribution of legitimate gradients $\mathbf g_h$, then generate its malicious update $\mathbf g_m$ to match that distribution. Because the malicious update looks normal, it can slip past many robust aggregation rules designed to reject outliers. The trade-off is that the attacker must refrain from making large deviations, so the impact on the global model—while harmful—may be gradual.

\textbf{Label Flipping (LF)~\cite{jebreel2022defendinglabelflippingattackfederated}:} In a label flipping attack, adversaries poison their local training data by relabeling examples from one class as another class. For instance, if there are $M_\ell$ classes, a client could transform labels $\ell$ to $M_\ell - \ell$ (effectively inverting the label encoding). The malicious client then trains on this mislabeled data and sends the resulting model updates to the server. Label flipping causes the global model to misclassify certain classes (those targeted by the flip) without necessarily triggering anomaly detection in gradients, since the gradient magnitudes and variances may appear normal.

\textbf{Data Injection (DI)~\cite{shalom2024data,shalom2022localization}:} Data injection attacks go one step further by adding fake or misleading data points to a client's local dataset. Unlike gradient manipulation, data injection affects the model indirectly via training. A malicious client can introduce out-of-distribution samples or engineered inputs that cause its local model update to skew the global model. Because the attack happens at the data level, the submitted gradient might not look obviously anomalous (especially if the client labels the injected data consistently with its attack goal). This makes pure gradient-based defenses less effective against DI.

While the above attacks assume the central server is honest and only clients are compromised, one can also consider the scenario of a \emph{malicious server}. A corrupted server has far greater power, since it can modify the aggregation of all client updates. One example of such an attack is \textbf{Inner Product Manipulation (IPM)} \cite{cao2019understanding}. In IPM, the attacker is the server itself (or an intruder who controls the server) and it alters the aggregated update to cancel out honest contributions. For instance, the server can compute the average of the honest gradients and then invert it with a scaling factor:
\begin{align}
    \mathbf g_m = -\,\epsilon \cdot \frac{\sum_{i \in V_h}\mathbf g_i}{|V_h|}, 
\end{align}
where $V_h$ is the set of honest clients and $\epsilon$ sets the attack strength. This manipulated "update" is then applied as if it were a normal aggregated gradient, effectively driving the global model in whatever direction the adversary chooses. Since all clients trust the server's aggregation, IPM can be devastating. Detection of such server-side attacks is difficult, but methods like BALANCE ~\cite{fang2024byzantine}  (discussed later in Section~7)  can sometimes spot its characteristic directional shifts.

\subsection{Backdoor attacks: preserving utility with hidden triggers}
Backdoor attacks implant an undetected “trigger” that, when present in a model’s input, induces malicious outputs. Crucially, the model remains accurate on ordinary inputs, making these attacks difficult to detect. Backdoors serve two main purposes: leaking private information (privacy-oriented backdoors) or altering predictions for the attacker’s benefit (model misbehavior).

\subsubsection{Backdoors targeting privacy breaches}
By actively modifying the training process, an attacker can embed triggers that force the model to reveal information about other clients. This can augment passive privacy attacks (Section~\ref{sec.passive}). For example, URVFL~\cite{yao2024urvfl} leverages a pre-trained encoder-decoder and an auxiliary classifier in vertical FL, inserting triggers that correlate model outputs with certain private features. Even defenses like differential privacy struggle to detect such refined backdoors.

\subsubsection{Backdoors for adversarial advantage}
Many backdoors aim to misclassify triggered inputs to a specific label without impairing overall performance. A single universal trigger (e.g., a small watermark in images) exemplifies a single-trigger backdoor \cite{bagdasaryan2020backdoor,9806416}, while multi-trigger backdoors\cite{xie2019dba,9806416} embed multiple, distinct triggers across different malicious clients. Multi-trigger approaches are more robust since discovering and neutralizing one trigger may leave others intact.

Some backdoor methods operate in a black-box setting, where the attacker sees only model outputs, yet can still iteratively adjust triggers by querying predictions \cite{zhang2020backdoor}. Others are \emph{adaptive} \cite{fang2020local}, dynamically modifying the poisoning rate or trigger shape in response to changing aggregator defenses. Since backdoors generally preserve high accuracy on benign inputs, they can remain hidden until the attacker provides an input with the correct trigger pattern.

\section{FL against active adversary II: defenses}\label{sec.robust}
Modern FL systems employ various defense strategies to counter active adversaries in both CFL and DFL settings. These approaches primarily focus on robust aggregation mechanisms that withstand malicious updates while preserving model utility. The following subsections analyze prominent defense methodologies and their operational characteristics, see the overall summery in Table \ref{tab:defense}.    %The Block-chain structure gives another view to realize defense. Our research mainly focuses on non-block-chain-based DFL, thus the concerned defense method will not be discussed in this section. 
\subsection{CFL defense approaches}
In CFL, the server orchestrates aggregation and  three principal robust aggregation techniques have demonstrated effectiveness:

\textbf{Trimmed Mean (TM)}~\cite{yin2018byzantine} operates through dimension-wise outlier removal. For each parameter dimension, the algorithm sorts client gradients $\{\mathbf g_i\}_{i=1}^n$, discards the $2k$ most extreme values ($k > |\mathcal{V}_b|$ where $\mathcal{V}_b$ represents Byzantine clients), and computes:
\begin{equation}
    \mathbf{g}_{\text{TM}} = \frac{1}{n-2k} \sum_{i \in \mathcal{V}\setminus\mathcal{V}_e} \mathbf{g}_i
    \label{eq:trimmed_mean}
\end{equation}
where $\mathcal{V}_e$ contains nodes with extreme gradient values. TM proves effective against moderate numbers of adversaries producing outlier updates.

\textbf{Median Aggregation (MED)}~\cite{yin2018byzantine,ye2024tradeoff} replaces mean aggregation with dimension-wise median selection. This non-parametric approach neutralizes the influence of extreme values without requiring prior knowledge of adversary counts:
\begin{equation}
    \mathbf{g}_{\text{MED}} = \text{median}(\{\mathbf{g}_i\}_{i=1}^n)
\end{equation}

\textbf{Krum Selection}~\cite{blanchard2017machine} implements a consensus-based approach by selecting the gradient vector closest to its neighbors through Euclidean distance analysis:
\begin{equation}
    \mathbf{g}_{\text{Krum}} = \mathbf{g}_{\arg\min_i \sum_{j \in \mathcal{S}_i} \|\mathbf{g}_i - \mathbf{g}_j\|^2}
    \label{eq:krum}
\end{equation}
where $\mathcal{S}_i$ contains the $n-|\mathcal{V}_b|-2$ nearest neighbors of client $i$. Multi-Krum variants extend this concept by aggregating multiple reliable gradients.

\subsubsection{Hierarchical Defenses}
The \textbf{Bulyan} framework~\cite{mhamdi2018hidden} combines Krum and TM through two-phase filtering: First, it selects candidate gradients via Krum-like consensus, then applies TM aggregation on this subset. This layered approach enhances robustness against sophisticated attacks attempting to bypass single-stage defenses.

\subsection{Decentralized FL Defenses}
\subsubsection{Local Inference Detection}
The \textbf{Local Injection Detection (LID)} method~\cite{shalom2022localization} implements dynamic neighbor screening through hypothesis testing. Each node $i$ evaluates updates from neighbor $j$ using:
\begin{equation}
    \Delta U_{j} = \frac{2}{\Delta T} \sum_{t=1}^{\Delta T/2} \| \mathbf{u}_{j}^{(2t)} - \text{median}(\{\mathbf{u}_{l}^{(2t)}\}_{l \in \mathcal{N}_i\setminus j}) \|_\infty
\end{equation}
where the infinity norm $\|\cdot\|_\infty$ captures maximum parameter deviations and $\mathbf{u}_{j}^{(t)}=\frac{1}{\eta^{(t)}}(\mathbf{w}_{j}^{(t+1)}-\mathbf{w}_{j}^{(t)})$, $\eta^{(t)}$ is time-varying step size. Updates are classified as malicious ($H_1$) or benign ($H_0$) through threshold comparison:
\begin{equation}
    \Delta U_{i,j} \underset{H_0^{i,j}}{\overset{H_1^{i,j}}{\lessgtr}} \delta_u \sqrt{|\mathcal{V}_i|}
\end{equation}
%The even-indexed time steps (2t) ensure update independence for reliable detection.

\subsubsection{Adaptive Filtering Approaches}
The \textbf{Uniform Byzantine-Resilient Aggregation Rule (UBAR)}~\cite{guo2022byzantine} employs dual-stage filtering: First, nodes select neighbors with minimal model distance; second, they validate updates through local performance evaluation. This combination ensures both geometric consistency and functional utility of aggregated models.

\subsubsection{Topology-Aware Defenses}
\textbf{Self-Centered Clipping (SCClip)}~\cite{he2023selfcentered} addresses non-IID challenges through localized update constraints:
\begin{equation}
    \mathbf{w}_i^{(t+1)} = \sum_{j\in\mathcal{N}_i} A_{ij} \left( \mathbf{w}_i^{(t)} + \text{Clip}(\mathbf{w}_j^{(t)} - \mathbf{w}_i^{(t)}, \tau_i) \right)
\end{equation}
where the clipping operation $\text{Clip}(\mathbf{x}, \tau) = \min(1, \tau/\|\mathbf{x}\|) \cdot \mathbf{x}$ bounds update magnitudes relative to local model $\mathbf{w}_i^{(t)}$.

\subsubsection{Dynamic Threshold Mechanisms}
The \textbf{Balance} method~\cite{fang2024byzantine} implements adaptive filtering through:
\begin{equation}
    \mathcal{S}_i^{(t)} = \left\{ j \in \mathcal{N}_i : \|\mathbf{w}_j^{(t)} - \mathbf{w}_i^{(t)}\| \leq \gamma^{(t)} \right\}
\end{equation}
where threshold $\gamma^{(t)}$ follows exponential decay $\gamma^{(t)} = \gamma_0 \cdot \alpha^t$ ($\alpha \in (0,1)$). This progressive tightening ensures increasing selectivity during training convergence.

\begin{table}[H]
    \centering
    \renewcommand{\arraystretch}{1.2}
    \scriptsize
    \begin{tabular}{l c c c c c c c}
        \toprule
        \textbf{Defense Approach} & \textbf{Assumed Topology} & \textbf{General Applicable} & \multicolumn{3}{c}{\textbf{Attacks}} & \multicolumn{2}{c}{\textbf{FL Formulation}}\\
        \cmidrule(lr){4-6} \cmidrule(lr){7-8}
        & & & \textbf{Explicit} & \textbf{Backdoor} & \textbf{Passive} & \textbf{Separate aggregation} & \textbf{Joint-optimization}\\
        \midrule
        \multicolumn{6}{l}{\textbf{Robust Aggregation}} \\
        \midrule
        TM\cite{yin2018byzantine} & CFL & $\checkmark$ & $\checkmark$ & $\times$ & $\times$ & $\checkmark$  & \\
        MED\cite{yin2018byzantine,ye2024tradeoff} & CFL & $\checkmark$ & $\checkmark$ & $\times$ & $\times$ & $\checkmark$  & \\
        Krum\cite{10.1145/3616390.3618283,blanchard2017machine} & CFL & $\times$ & $\checkmark$ & $\times$ & $\times$ &$\checkmark$  & \\
        Bulyan\cite{mhamdi2018hidden} & CFL & $\times$ & $\checkmark$ & $\times$ & $\times$ & $\checkmark$ & \\
        LID\cite{shalom2022localization} & DFL & $\times$ & $\checkmark$ & $\checkmark$ & $\times$ & $\checkmark$ & \\
        UBAR\cite{guo2022byzantine} & DFL & $\times$ & $\checkmark$ & $\checkmark$ & $\times$ &  & $\checkmark$\\
        SCClip\cite{he2023selfcentered} & DFL & $\times$ & $\checkmark$ & $\checkmark$ & $\checkmark$ &  & $\checkmark$ \\
        Balance\cite{fang2024byzantine} & DFL & $\times$ & $\checkmark$ & $\checkmark$ & $\checkmark$ &  & $\checkmark$\\
        % \midrule
        % Quantum FL\cite{xia2021defending,zhang2022secure,ren2023quantumfl} & CFL & $\checkmark$ & $\checkmark$ & $\times$ & $\times$ &  & \\
        % \midrule
        % Differential Privacy\cite{9069945,ye2024tradeoff} & CFL/DFL & $\checkmark$ & $\checkmark$ & $\times$ & $\times$ & $\checkmark$ & $\checkmark$\\
        \bottomrule
    \end{tabular}
    \caption{Comparison of defense approaches against different types of attacks.}
    \label{tab:defense}
\end{table}

Overall, relies on server authority and global statistical analysis, prioritizing scalability at the cost of introducing a single point of failure. In contrast, DFL distributes trust through localized validation and outlier filtering mechanisms, which better adapt to dynamic environments but complicate model convergence in heterogeneous data settings.
% \subsection{Discussions}

\section{Conclusion}\label{sec.conclusion}
This survey reframes the evolution of FL by arguing that the core distinction between CFL and DFL lies not only in topology but also in protocol design: separated aggregation versus joint optimization. Our taxonomy reveals critical insights: (1) Most DFL methods adopt heuristic averaging protocols rather than rigorous distributed optimization, limiting their robustness and scalability; (2) Protocol choice directly governs privacy-utility trade-offs, with joint optimization methods showing untapped potential for privacy preservation through inherent constraints.  

Key gaps persist, particularly in DFL’s underuse of distributed optimization frameworks despite their theoretical advantages for non-IID data and Byzantine resilience. Future work should prioritize protocol-aware designs that unify optimization and robustness, adaptive defenses against evolving attacks, and hybrid architectures balancing CFL’s efficiency with DFL’s security. By transcending topology-centric paradigms, the field can unlock FL’s full potential as a scalable, private, and attack-resilient framework for decentralized collaboration.

\bibliographystyle{unsrt}
% \bibliography{dualpath.bib} 

\end{document}